\documentclass[a4paper,conference]{IEEEtran}
\usepackage{algpseudocode}
\usepackage{algorithm}
 \usepackage{multirow}
\usepackage{booktabs}
\usepackage[utf8]{inputenc}
\usepackage{times}
\usepackage{epsfig}
\usepackage{graphicx}
\usepackage{amsmath}
\usepackage{amssymb}
\usepackage{wrapfig}
\usepackage{xcolor}

\ifCLASSINFOpdf
\else
\fi

\ifCLASSOPTIONcompsoc
 \usepackage[caption=false,font=normalsize,labelfont=sf,textfont=sf]{subfig}
\else
 \usepackage[caption=false,font=footnotesize]{subfig}
\fi
\usepackage{fixltx2e}
\usepackage{stfloats}
\usepackage{url}
\begin{document}
\title{Swapping Semantic Contents for Mixing Images}

% \author{\IEEEauthorblockN{Rémy Sun}
% \IEEEauthorblockA{School of Electrical and\\Computer Engineering\\
% Georgia Institute of Technology\\
% Atlanta, Georgia 30332--0250\\
% Email: http://www.michaelshell.org/contact.html}
% \and
% \IEEEauthorblockN{Homer Simpson}
% \IEEEauthorblockA{Twentieth Century Fox\\
% Email: homer@thesimpsons.com}
% \and
% \IEEEauthorblockN{James Kirk\\ and Montgomery Scott}
% \IEEEauthorblockA{Starfleet Academy\\
% San Francisco, California 96678--2391\\
% Telephone: (800) 555--1212\\


\author{\IEEEauthorblockN{Rémy Sun\IEEEauthorrefmark{1,2},
Clément Masson\IEEEauthorrefmark{2},
Gilles Hénaff\IEEEauthorrefmark{2},
Nicolas Thome\IEEEauthorrefmark{3} and
Matthieu Cord\IEEEauthorrefmark{1}}
\IEEEauthorblockA{\IEEEauthorrefmark{1}MLIA, ISIR, Sorbonne Université, Paris, France}
\IEEEauthorblockA{\IEEEauthorrefmark{2}Thales Land and Air Systems, Elancourt, France}
\IEEEauthorblockA{\IEEEauthorrefmark{3}VERTIGO, CEDRIC, Conservatoire National des Arts
  et Métiers, Paris, France}}


\maketitle



% \fi
\IEEEpeerreviewmaketitle

\begin{abstract}
 Deep architecture have proven capable of solving many tasks provided a
sufficient amount of labeled data. In fact, the amount of available labeled data
has become the principal bottleneck in low label settings such as
Semi-Supervised Learning. Mixing Data Augmentations do not typically yield new
labeled samples, as indiscriminately mixing contents creates between-class
samples. In this work, we introduce the SciMix framework that can learn to
replace the global semantic content from one sample. By teaching a StyleGan
generator to embed a semantic style code into image backgrounds, we obtain new
mixing scheme for data augmentation. We then demonstrate that SciMix yields
novel mixed samples that inherit many characteristics from their non-semantic
parents. Afterwards, we verify those samples can be used to improve the performance
semi-supervised frameworks like Mean Teacher or Fixmatch, and even fully
supervised learning on a small labeled dataset.
\end{abstract}

%%%%%%%%% BODY TEXT
\section{Introduction}

Deep architectures have proven capable of reliably solving a variety of tasks
such as classification \cite{he16_ident_mappin_deep_resid_networ,
krizhevsky2012imagenet}, object detection \cite{ren2015faster} or machine
translation \cite{vaswani2017attention}. This is however contingent on there
being a large amount of labeled data to train models on. This is seldom the case
in practical applications where labelisation tends to be costly.

% Data Augmentation addresses this issue by artificially inflating the
% amount of available data. Indeed, it has long been known that applying
% perturbations to samples allows models to train on smaller training sets
% \cite{test}. % While the perturbations applied have historically remained fairly small
% % \cite{tes}, the last few years have seen the emergence of stronger transformations \cite{test}.
% % These stronger augmentation techniques demonstrate that models benefit from training on
% % samples that significantly differ from real samples.
% At its core, data augmentation succeeds because it postulates even strong
% transformations preserve the semantic content of inputs.

% Mixing data augmentation techniques like MixUp \cite{test} challenge this by
% combining multiple samples. Since those samples can have different semantic
% content, the generated hybrids cannot teach the model to be invariant to the
% applied perturbations. Indeed, these augmentation techniques \cite{cite}
% typically induce models to predict interpolated labels on the generated hybrids.
% While this has been proven to successfully regularize models, this new family of generative
% data augmentations loses much of the original purposes of data augmentations

Data Augmentation \cite{he19_data_augmen_revis, he18_bag_trick_image_class_with} - the creation of artificial samples from existing ones - has
long been used to help models train on small datasets. Of particular
interest in low label settings like Semi-Supervised Learning \cite{ChaSchZie06, berthelot19_mixmat} (SSL), Mixing Samples Data
Augmentations \cite{zhang2017mixup, yun2019cutmix} (MSDA)
can be used to combine the few samples that are either labeled or reliably
pseudo-labeled with the large pool of unlabeled data. Unfortunately, Mixing Data
Augmentations mix contents indiscriminately and as such create between-class
hybrids for classification. While such hybrids have proven very useful for model regularization
\cite{carratino2020mixup, thulasidasan2019mixup},
this process strongly perturbs the semantic information from reliably labeled samples.

\begin{figure}
  \centering
  \includegraphics[width=.8\linewidth]{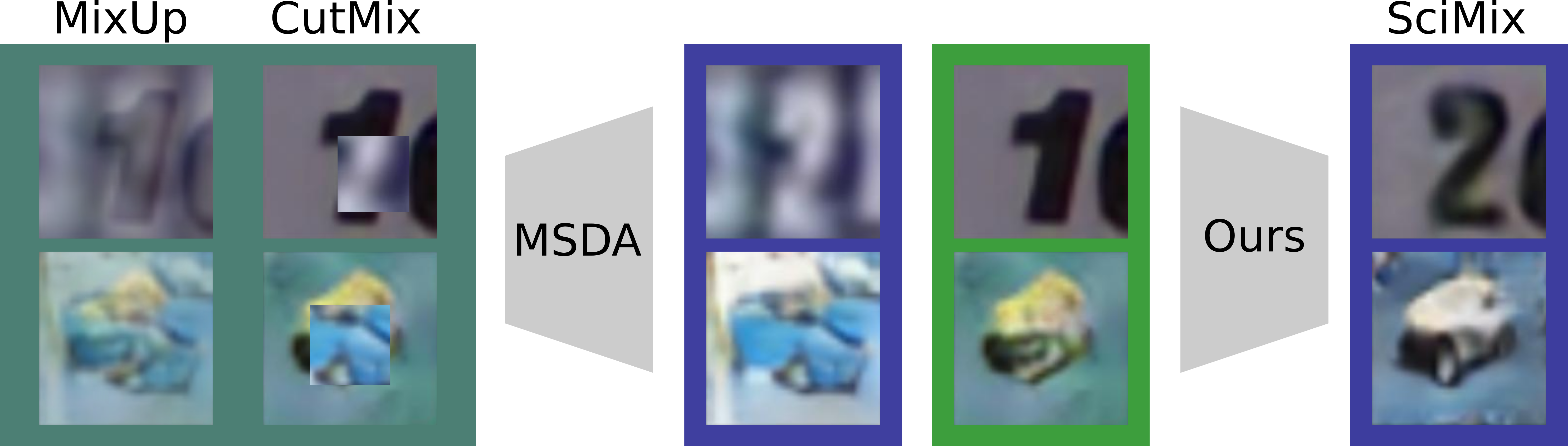} 
  \caption{Standard mixing augmentations mix contents indiscriminately
    whereas our method mixes the semantic number from one sample with the
    background information from another.}
  \label{fig:intro}
\end{figure}

We argue that, with the right adjustments, mixing data augmentations can still be used to teach
semantic invariance to neural networks. Indeed, if we can mix the semantic content of one sample
with the non-semantic content of another, then the generated samples will still be
actual in-class samples. For instance, Fig.~\ref{fig:intro}
shows that mixing two street numbers with MixUp or CutMix typically leads to no real number appearing
on the image whereas carefully selecting the contents to be mixed leads to a
mixed sample that remains realistic.

In this paper, we introduce SciMix a new framework that learns to separate semantic
from non semantic content and generate hybrids that preserve most of the
specified non semantic content while still properly representing the required
semantic content. Moreover, we demonstrate hybrids generated from this framework
improve model performance through extensive low label experiments, primarily on the
Semi-Supervised Learning problem (CIFAR10 and SVHN).
We therefore propose three main contributions:
\textbf{1)} A new mixing paradigm designed to create artificial in-distribution samples that
  embed the non-semantic content of one sample into the non-semantic context of
  another. This new approach generates a new type of data
  augmentation for deep learning.
\textbf{2)} A new auto-encoding architecture and associated learning scheme
  that trains a generator to mix semantic and non-semantic
  contents. In particular, we purposefully train a model to separate
  semantic and non-semantic contents into two representations, and train a
  style-inspired generator to embed the semantic content (``style'' code) into the
  non-semantic background (traditional input).
\textbf{3)} A new learning process to leverage our new mixing data augmentation. We
  show mixed samples can be used to optimize an additional supervised objective
  that significantly improves classifier performance.

% We then demonstrate the benefits of our approach for
% Semi-Supervised Learning on CIFAR10 and SVHN through extensive experiments and
% ablation studies, and provide a preliminary fully supervised.

\section{SciMix Framework}

We propose in this paper a new mixing data augmentation that mixes the semantic
content of a sample with the non-semantic content of another. As such, we first
detail our scheme to train a model capable of mixing samples as per our exact
specifications (Sec.~\ref{sec:gen}). We then explain our data augmentation
strategy for visual classification leveraging our generated semantic hybrid
samples (Sec.~\ref{sec:da}). Finally, we discuss SciMix in the broader
context of content mixing techniques (Sec.~\ref{sec:rw}).

\subsection{Learning to generate hybrids}
\label{sec:gen}

%We start by designing an auto-encoding model that can learn to generate hybrids mixing the semantic content of one sample and the non-semantic content of another. The auto-encoder makes use of five neural networks: a semantic encoder $E_c$, a non-semantic encoder $E_r$, a generator/decoder $G$, a linear classifier $C$ and a discriminator $D$. 

\begin{figure*}
  \subfloat[Semantic auto-encoder of SciMix's generator. $x$ is
  encoded into $z_c$ (made semantic by $\mathcal{L}_S$) and $z_r$, which are decoded into $\hat{x}$
  optimized by $\mathcal{L}_{AE}=\mathcal{L}_{rec}+\mathcal{L}_{adv,r}$.\label{fig:ae}]{\includegraphics[width=.45\linewidth]{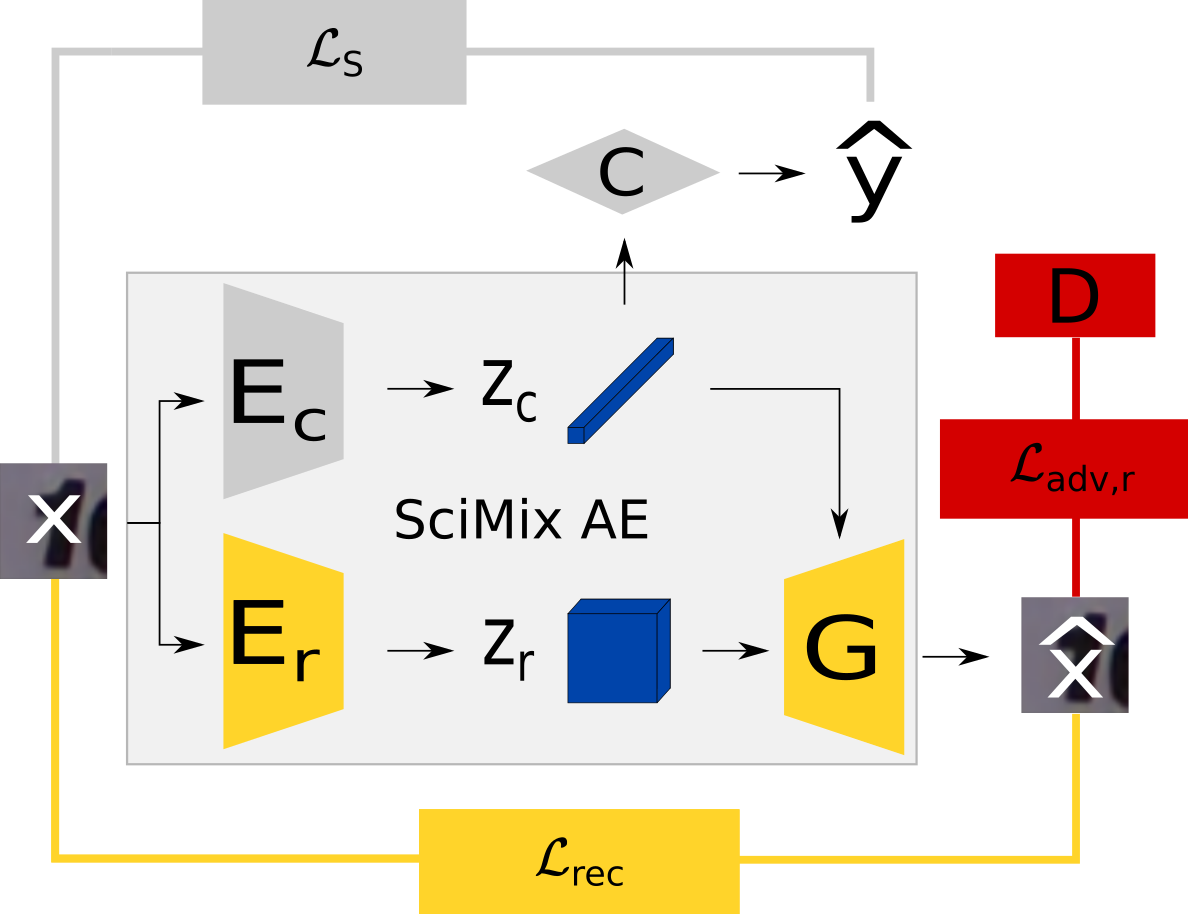}}
  \hfill
  \subfloat[Hybridization in SciMix's generator. Semantic components are
  extracted from $x_1$ and $x_2$ to obtain a hybrid optimized through
  $\mathcal{L}_{hyb}$ (detailed in 5 sub-losses here).\label{fig:hyb}]{\includegraphics[width=.45\linewidth]{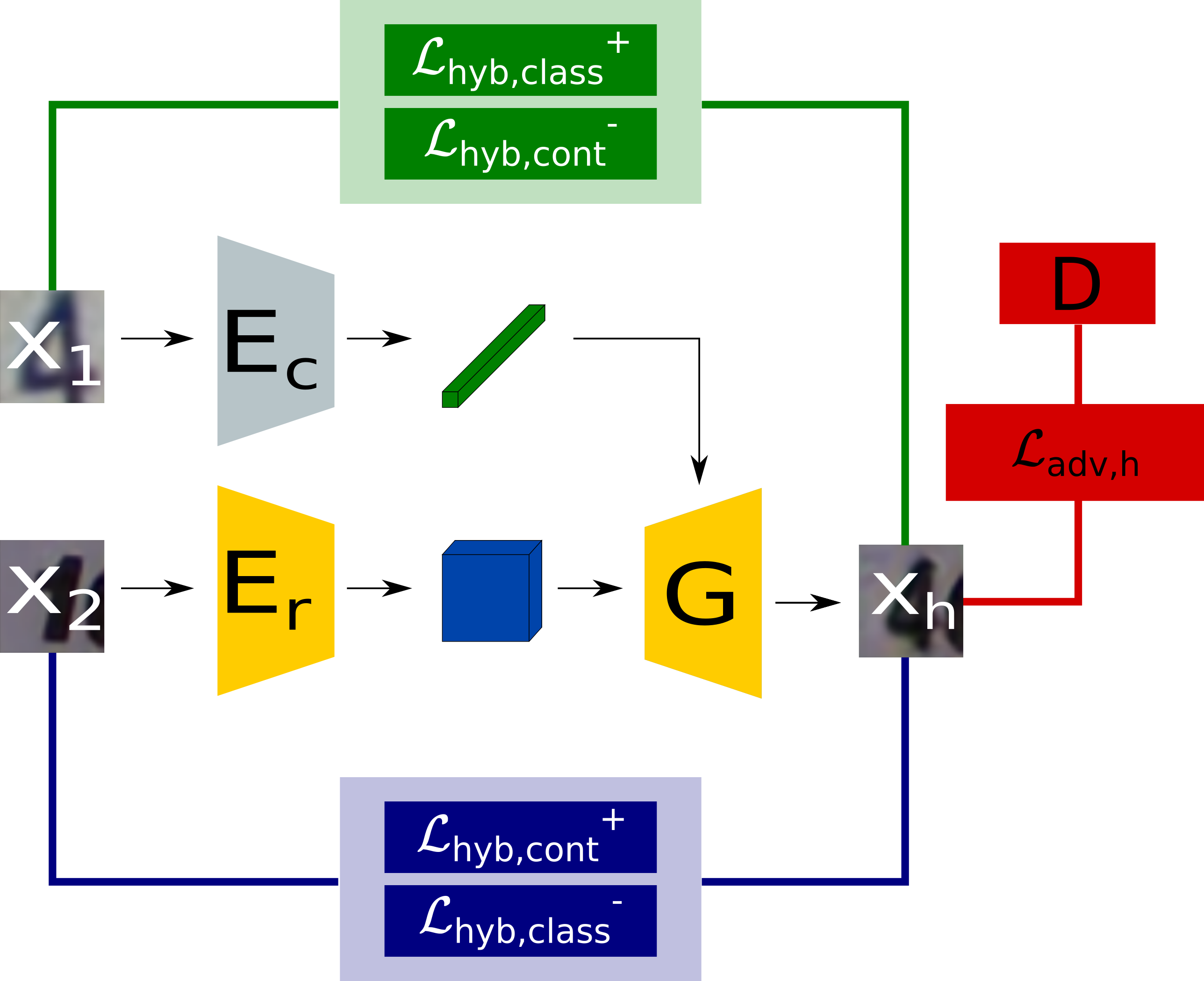}}
  \caption{Overview of the SciMix generator architecture. SciMix
    trains an auto-encoder with two latent spaces, one of which is semantic}
  \label{fig:overview}
\end{figure*}

\paragraph{Auto-encoding architecture} 
Our framework is based on a novel auto-encoder architecture presented in
Fig.~\ref{fig:ae} that treats semantic information as a global characteristic of an encoded image.
An input $x$ is projected into a semantic latent space $z_c$ by an
encoder $E_c$ as well as a complementary non-semantic latent space $z_r$ by an
encoder $E_r$. While our framework should primarily be understood as an
auto-encoder framework, the distinct nature of the latent spaces $z_c$ and $z_r$
requires more careful consideration. We choose in this paper to focus on the
definition and exploitation of the semantic features $z_c$, and simply treat
$z_r$ as information irrelevant to $z_c$. In other words, we design the
framework so that the semantic information $z_c$ controls what the generator $G$
reconstructs. As the notion of semantic information is fundamentally tied to
that of the tasks under consideration, we define $z_c$ with respects to a
classifier. More precisely, we treat the semantic latent space $z_c$ as the
feature space of a classifier. To this end, we add a linear neural layer $C$ on
top of $z_c$ (see Fig.~\ref{fig:ae}) that outputs a class prediction $\hat{y}$.
Note that $E_c \circ C$ therefore constitutes a standard CNN classifier
\cite{he16_ident_mappin_deep_resid_networ,
    zagoruyko2016wide}. We ensure the classifier $E_c \circ C$ correctly learns semantic
information through classification loss term $\mathcal{L}_{S}$ on its
output \footnote{$\mathcal{L}_S$ can correspond to any classifier training framework
  (e.g. supervised
  training, FixMatch \cite{sohn20_fixmat}). In semi-supervised experiments, we use Mean Teacher
  \cite{tarvainen17_mean_teach_are_better_role_model} as a simple classifier guide in order to leverage SSL datasets.}.

% \footnote{In this paper, we use Mean Teacher
%   \cite{tarvainen17_mean_teach_are_better_role_model} as a classifier guide in order to leverage Semi-Supervised datasets.
%   $\mathcal{L}_S$ could however be chosen as any kind of classifier training framework
%   (e.g. supervised
%   training, FixMatch \cite{sohn20_fixmat}).}.

Contrarily to \cite{NEURIPS2020_50905d7b}, we complete the encoding with a generator G that
computes a reconstruction $\hat{x}$ using the non-semantic features $z_r$ as
direct inputs and the semantic features $z_c$ as style codes. This makes the
non-semantic content $z_r$ easier to transfer, which is fortunate considering we
can ensure semantic transfer more easily through a classifier $E_c \circ C$ (see
Sec.~\ref{sec:variants} for the converse approach). As we treat the
model as an autoencoder, we use a reconstruction loss $\mathcal{L}_{rec}$ to
tie the reconstruction $\hat{x}$ to the input $x$. We further refine this
reconstruction by using an adversarial critic to smooth out details \cite{NEURIPS2020_50905d7b}.
We add a discriminator network $D$ (see Fig.~\ref{fig:ae}) to predict whether
the considered image is a real sample or a reconstruction . Conversely, $E_r$,
$E_c$ and $G$ are trained to fool $D$ into
seeing reconstructions $\hat{x}$ as real images. The resulting loss
$\mathcal{L}_{adv,r}$ serves to improve reconstructions learned through
the reconstruction loss.

\begin{equation}
  \label{eq:ae}
  \begin{split}
    \mathcal{L}_{AE} = \mathcal{L}_{rec}&+\mathcal{L}_{adv,r} = \sum_{x \in \mathcal{D}} \lVert x - G(E_c(x), E_r(x)) \rVert_2 \\
    & + \sum_{x \in \mathcal{D}} - \log(D(G(E_c(x), E_r(x)))).
  \end{split}
\end{equation}

Finally, our learning scheme is based on the minimization of the loss $\mathcal{L}_{gen}$ composed of $\mathcal{L}_{S}$ and $\mathcal{L}_{AE}$, plus an additional loss $\mathcal{L}_{hyb}$ term:

\begin{equation}
  \label{eq: main}
  \mathcal{L}_{gen} = \mathcal{L}_{AE} + \mathcal{L}_{S} + \mathcal{L}_{hyb},
\end{equation}

which is described in the following hybridizing
scheme.

\paragraph{Hybridization losses}

Simply training the auto-encoder architecture, even with $z_c$ as the feature
space of a classifier, is not enough to ensure that hybrids correctly inherit
characteristics from their parents. To force the model to properly inject
semantic content into the general background of known samples, we design
explicit hybridization losses (studied more closely in Sec.~\ref{sec:variants})
\begin{equation}
  \begin{split}
    \mathcal{L}_{hyb} = &\mathcal{L}_{hyb,class}^+ +
    \mathcal{L}_{hyb,cont}^+ \\ &+ \mathcal{L}_{hyb,class}^- + \mathcal{L}_{hyb,cont}^-
    + \mathcal{L}_{adv,h}.
  \end{split}
\end{equation}

% \paragraph{Semantic hybridization criterion}

$\mathcal{L}_{hyb,class}^+$ explicitly trains our model to rely on $z_c$ to
generate the main semantic object in the generated reconstruction/hybrid.
Indeed, we rely on the classifier $E_c \circ C$'s ability to identify and
classify the main object in inputs (see Fig.~\ref{fig:hyb}). 

Put plainly, we generate hybrids $x_h=G(E_c(x_1), E_r(x_2))$ from pairs of samples
in a batch and obtain logits predictions $C(E_c(x_h))$ for those hybrids.
$\mathcal{L}_{hyb, class}^+$ optimizes the model so that this prediction on the
logits match the prediction on the semantic parent $x_1$ of $x_h $. Importantly,
we only optimize the hybridization process that generates $x_h$: we do not
optimize the classifier's prediction on $x_h$ or $x_1$. The idea is that the autoencoder
learns to place $x_h$ in the right class manifold, while said class manifold
does not move to accommodate $x_h$:

\begin{equation}
  \label{eq:hyb_class}
  \begin{split}
    \mathcal{L}_{hyb, class}^+ = \sum_{x \in \mathcal{D}} \lVert &C(E_c(x_1)) \\&- C(E_c(G(E_c(x_1), E_r(x_2)))) \rVert_2.
  \end{split}
\end{equation}

% \paragraph{Non-semantic hybridization criterion}

Similarly, $\mathcal{L}_{hyb,cont}^+$ optimizes the model so that a generated
hybrids $x_h$'s non-semantic representation $z_{r,h}$ matches its non semantic
parent $x_2$'s non semantic component $z_{r,2}$. As in the semantic case, we
only optimize the generative process that leads to the generation of $x_h$ but do
not optimize $E_r$ to project $x_h$ close to its non-semantic parent:

\begin{equation}
  \mathcal{L}_{hyb, cont}^+ = \sum_{x \in \mathcal{D}} \lVert E_r(x_2) - E_r(G(E_c(x_1), E_r(x_2))) \rVert_2.
\end{equation}

% \paragraph{Orthogonalization criterion}

We also train hybrids to differ from their parents through the negative semantic hybridization
loss $\mathcal{L}_{hyb, class}^- = \sum_{x \in \mathcal{D}} - \lVert
C(E_c(x_2)) - C(E_c(G(E_c(x_1), E_r(x_2)))) \rVert_2$ and the negative non-semantic
hybridization loss $\mathcal{L}_{hyb,
cont}^-=\sum_{x \in \mathcal{D}} - \lVert E_r(x_1) - E_r(G(E_c(x_1), E_c(x_2)))
\rVert_2$. In practice, this means maximizing the distance between hybrids and
semantic (resp. non-semantic) parent in non-semantic (resp. semantic) space.

% \paragraph{Quality control through adversarial optimization}

To ensure the quality of generated hybrids, we train the discriminator $D$ to also recognize hybrids as synthetic
images. With this discriminator we can simply add an adversarial loss term
$\mathcal{L}_{adv, h}$ to ensure hybrids look realistic (as far as $D$ is concerned).

% ------------------------------------------------------------------------

\subsection{Training a classifier by leveraging our Data Augmentation}
\label{sec:da}

We now have a novel mixing data augmentation that can embed the semantic content
of one sample to the non-semantic context of other samples, given a trained
generator. This provides a useful and new way to improve any standard training
method ``X'' by adding a single additional loss term $\mathcal{L}_{contradict}$: $\mathcal{L}_{SciMix} =
\mathcal{L}_{X} + \mathcal{L}_{contradict}$.

\paragraph{Generating hybrids given a trained autoencoder}

Generating hybrids given a trained model is straightforward
(Fig.~\ref{fig:hyb} shows how a hybrid is mixed). Specifically, given samples $x^{(1)}$ (with known label $y^{(1)}$) and $x^{(2)}$, we
extract the relevant features $z_c^{(1)}=E_c(x^{(1)})$,
$z_r^{(1)}=E_r(x^{(1)})$, $z_c^{(2)}=E_c(x^{(2)})$ and $z_r^{(2)}=E_r(x^{(2)})$.
$x_h=G(z_c^{(1)},z_r^{(2)})$ is now a sample with class $y^{(1)}$. As a
conservative measure,
we only keep the generated hybrid if $C(E_c(x_h)) = y^{(1)}$
to avoid disturbing decision boundaries too much. Note that with this, we
generate a strong augmentation of $x_1$ and teach the classifier to group $x_1$
with its strongly augmented version in a similar line to work in contrastive
representation learning \cite{he19_momen_contr_unsup_visual_repres_learn}.

\paragraph{Training a new classifier $f$}

We now propose a way to leverage our novel hybrids to improve the training of
standard models such as Mean Teacher \cite{tarvainen17_mean_teach_are_better_role_model} or FixMatch \cite{sohn20_fixmat}. To
this end, we compute hybrids that mix the semantic content of each sample in 
the batch with the non-semantic content of other samples in the batch. We
leverage those hybrids by optimizing an additional loss:
\begin{equation}
  \begin{split}
    \mathcal{L}_{contradict} = \sum_{x_c, x_r \in \mathcal{B},perm(\mathcal{B})} &[ l_{MSE}(f(x_h),
      \alpha * f(x_c) \\ &+ (1-\alpha) f(x_r))].
  \end{split}
\end{equation}

This new loss takes advantage of our mixing paradigm
by mostly imputing the semantic parent's label to our hybrids, with only a
slight dependence on the non-semantic parent to acknowledge the imperfection of
the mixing process. Contrarily to standard mixing augmentations, the ratio
$\alpha>0.5$ is a fixed hyperparameter (in the spirit of label smoothing
\cite{smoothing}).

\subsection{Mixing contents in the literature}
\label{sec:rw}

Mixing of one type of content with another is more readily found in unsupervised
image-to-image translation: models are trained to translate the content of one
image to the ``domain'' of another, though these terms are rarely well defined.
Interestingly, bi-modal auto-encoding architectures appear fairly early on in
this literature \cite{NIPS2017_dc6a6489, Huang_2018_ECCV}. Incidentally, more
recent works in few-shot translation \cite{Liu_2019_ICCV} and unsupervised
translation \cite{baek2021rethinking} have even started associating the domain
(or class) information to a style code fed as input to a StyleGan inspired
decoder. In a more supervised fashion, such methods have been used to combine
textures (domain information) with structural information (image information)
\cite{NEURIPS2020_50905d7b}. This line of work however is specifically tailored
to image generation and fails to leverage information to from fully fledged
classifier to learn complex semantic variations.

Style transfer actually tackles the issue of mixing different types of contents
in a similar fashion to our framework. The main difference between such
frameworks and our problem lies in the definition of the contents to mix. In
style transfer, the distinction is made between a style code and a structure
code while we seek to mix semantic and non-semantic content. As such, our work
reprises feature map modulation mechanisms that have been proven to work in
style transfer \cite{huang17_arbit_style_trans_real_time, karras2019style,
stylegan3} but makes use of additional losses to ensure we mix semantic and
non-semantic contents.

It is worth noting that our goal of generating images in a way such that
semantic content can be modified independently of non semantic content echoes
that of disentangled generation \cite{higgins2017beta, dualdis,
  NIPS2016_45fbc6d3, Bousmalis_2017_CVPR}. Importantly however, we
aim to modify only a single coarse attribute rather than separate a multitude of
fine grained characteristics.

\section{Experiments}

We demonstrate here how our SciMix data augmentation can be leveraged to improve
training in low-label
settings. To this end, we conduct extensive experiments on the semi-supervised
problem on the CIFAR10 \cite{krizhevsky2009learning} and SVHN
\cite{netzer2011reading} datasets. Additionally, we propose in
Sec.~\ref{sec:cub} a study of SciMix's performance in a fully supervised setting
with few labels on a variation of the CUB-200 \cite{WelinderEtal2010} dataset.

In the semi-supervised case, we show SciMix improves two backbone methods:
Mean Teacher \cite{tarvainen17_mean_teach_are_better_role_model} and FixMatch
\cite{berthelot19_mixmat} (refer to Sec.~\ref{sec:da} for how we apply our framework
to these methods). We chose Mean Teacher as a
reference consistency-based baseline. Beyond its widespread use in SSL,
consistency induces a stabilization we feel would help
extract invariant semantic features. FixMatch is a state-of the art
SSL method based on strong augmentation, and often serves as a reference or backbone
in the literature \cite{Li_2021_ICCV, zhang2021flexmatch}.

We operate on a standard WideResNet-28-2
\cite{zagoruyko2016wide} for our classifiers (both $f$ and $E_c \circ C$).
$E_r$ follows the same architecture as $E_c$. The skeleton of $G$ follows
a StyleGanv2 \cite{Karras_2020_CVPR} architecture.
Hyperparameters and optimizers were generally taken to
follow settings reported in the base methods' original papers
\cite{tarvainen17_mean_teach_are_better_role_model, berthelot19_mixmat}.

We report the $mean\pm std$ classification accuracy over 3 seeded runs for varying numbers of
labeled samples in a dataset (the rest are treated as unlabeled). The
SciMix generators used to train a model with $N$ labeled samples are also
trained with only $N$ labeled samples. One generator is trained per setting, and
classifiers trained with the generator's mixed samples are trained on the same
split of labeled/unlabeled data to avoid information leakage. More details for all experiments are
given in Appendix.

\subsection{Performance gains}

\begin{table*}
  \caption{Mixing samples with SciMix as a data augmentation improves the
    performance of Mean Teacher and FixMatch. Significant accuracy (\%) gains
    are observed on CIFAR10 (with 100, 250 and 500 labels) and SVHN (with 60 and
    100 labels).}
  \label{tab:gains}
  \centering
  \subfloat[Mean Teacher\label{tab:mt}]{
  \resizebox{1.25\columnwidth}{!}{
  \begin{tabular}{lccccc}
    \toprule
    \multirow{2}{*}{Method} & \multicolumn{3}{c}{CIFAR10} & \multicolumn{2}{c}{SVHN} \\ \cmidrule(r){2-4}\cmidrule(r){5-6}
                            & 100 & 250 & 500 & 60 & 100 \\ \midrule
    % Purely supervised (lower bound) & & $27.8 \pm 0.9$ & $35.4 \pm 1.9$& &  \\ \midrule
    Mean Teacher, \cite{tarvainen17_mean_teach_are_better_role_model} & $40.5 \pm 6.4$ & $63.1 \pm 0.9$ & $72\pm 3$ & $48.7 \pm 23.0$ & $82.3 \pm 5.5$ \\
    SciMix w/ Mean Teacher & $\mathbf{46.4 \pm 1.2}$ & $\mathbf{68.0 \pm 1}$ & $\mathbf{77.2 \pm 0.5}$ & $\mathbf{83.4 \pm .5}$ & $\mathbf{87.3 \pm 2.8}$  \\ \bottomrule
  \end{tabular}
  }
  }
  \subfloat[FixMatch\label{tab:fix}]{
  \resizebox{.65\columnwidth}{!}{
  \begin{tabular}{lcc}
    \toprule
    \multirow{2}{*}{Method} & \multicolumn{1}{c}{CIFAR10} & \multicolumn{1}{c}{SVHN} \\ \cmidrule(r){2-2}\cmidrule(r){3-3}
                            & 100 & 60 \\ \midrule
    FixMatch, \cite{berthelot19_mixmat} & $88.6 \pm 0.7$ & $\mathbf{96.4 \pm 0.3}$  \\
    SciMix w/ FixMatch & $\mathbf{90.7 \pm 0.2}$ & $\mathbf{96.5 \pm 0.9}$ \\ \bottomrule
  \end{tabular}
  }
  }
\end{table*}

Tab.~\ref{tab:gains} shows that adding our optimization on hybrids with
$\mathcal{L}_{contradict}$ - as described in Sec.~\ref{sec:da} - does indeed lead to improved performance on CIFAR10
and SVHN. Indeed, training with SciMix hybrids leads to significant accuracy
gains with a Mean Teacher classifier with a wide range of labeled samples.
We also observe improvements over FixMatch on very low labeled settings
(FixMatch performance quickly saturates on higher labeled settings).
Interestingly, two concurrent behaviors can be observed on Mean Teacher: SciMix hybrids become
more useful when less labeled samples are available but the quality of generated
hybrids becomes unreliable with few labeled samples. Indeed, at 100 labeled
samples on CIFAR10, the Mean Teacher classifier used to train the generator is
too weak to provide very useful hybrids. With a strong SciMix hybridizer
(trained with all samples), SciMix data augmentation would bring a Mean Teacher
classifier to an accuracy of $60.4 \pm 1.7$ with 100 labels on CIFAR10.

\subsection{Quality of hybridization}
\label{sec:exh}

\begin{figure}
  \subfloat[Examples of SciMix hybrids.\label{fig:exh_sem}]{
    \begin{minipage}{.09\linewidth}
      {\includegraphics[width=\linewidth]{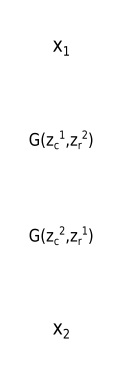}}
    \end{minipage}
    \begin{minipage}{.26\linewidth}
      {\includegraphics[width=\linewidth]{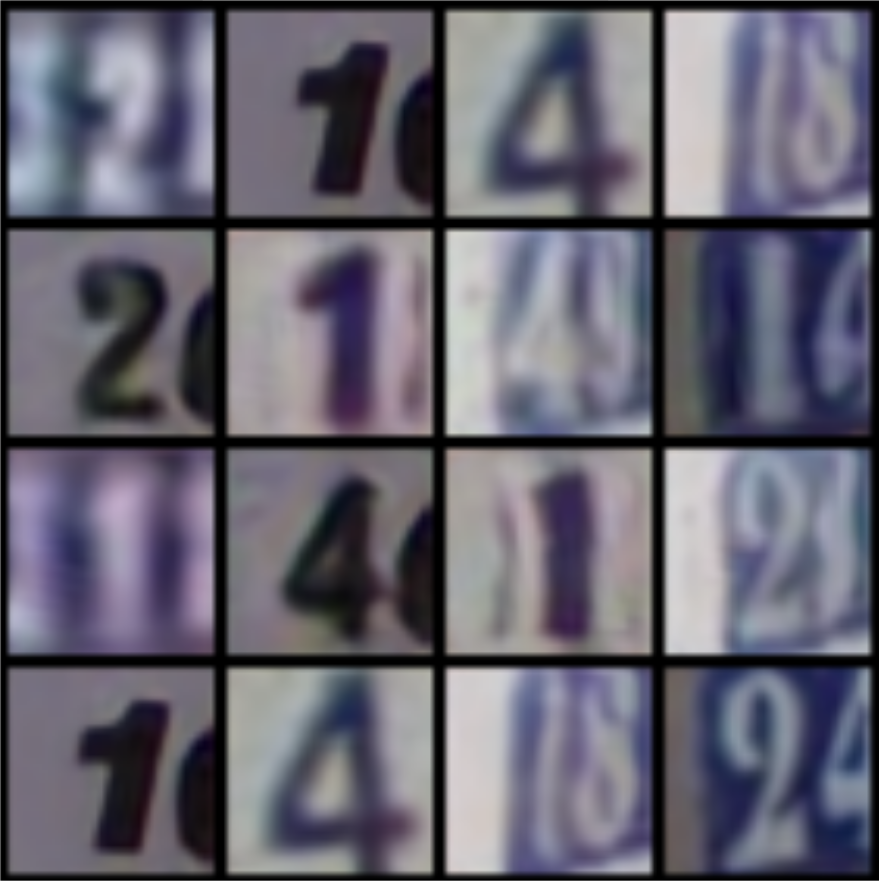}}
    \end{minipage}
    \begin{minipage}{.25\linewidth}
      {\includegraphics[width=\linewidth]{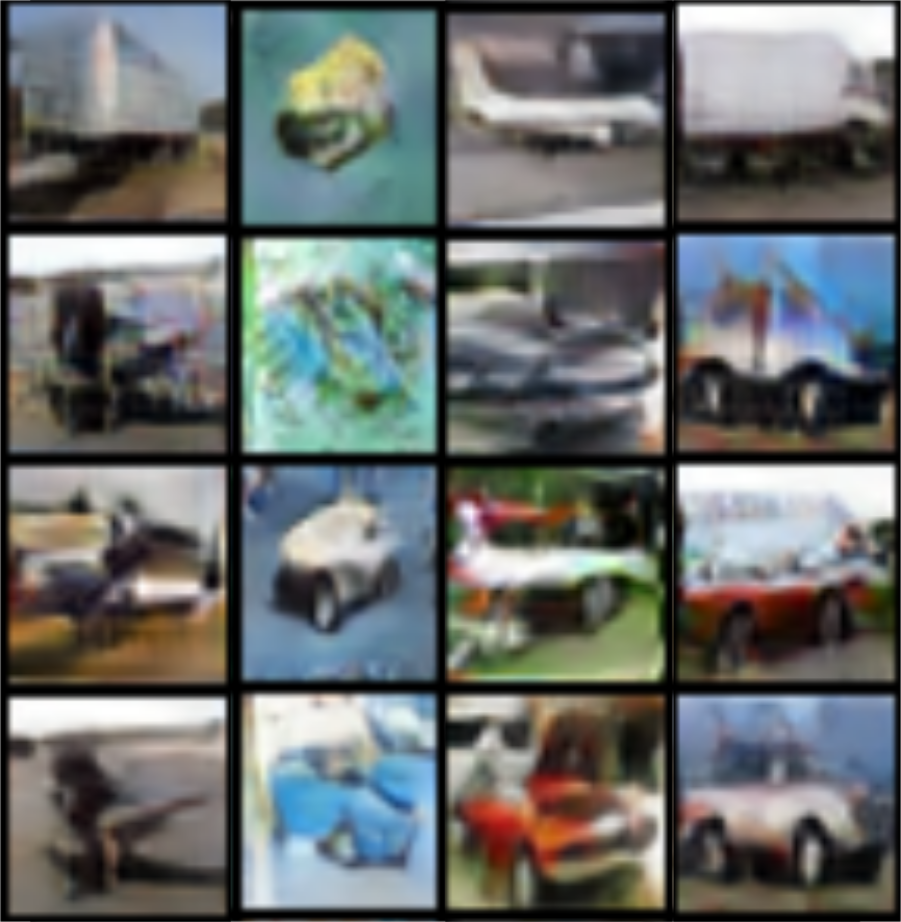}}
    \end{minipage}
  }
  \subfloat[Semantic (resp. non-semantic) transfer in hybrids.\label{tab:exh}]{
    \resizebox{.35\columnwidth}{!}{
  \begin{tabular}{lcc}
    \toprule
    Method & & \multicolumn{1}{c}{SVHN} \\ \midrule
                            & \multicolumn{1}{l}{FDA \cite{Yang_2020_CVPR}} &  \multicolumn{1}{c}{$77.0$}  \\
    Semantic           & \multicolumn{1}{l}{AdaIN \cite{huang17_arbit_style_trans_real_time}} &   \multicolumn{1}{c}{$92.1$}  \\
    transfer $s_c$              & MixUp \cite{zhang2017mixup,inoue18_data_augmen_by_pairin_sampl_images_class} & \multicolumn{1}{c}{$94.2$} \\
                            & \multicolumn{1}{l}{SciMix} & $\mathbf{95.2}$ \\ \midrule
                            & \multicolumn{1}{l}{FDA \cite{Yang_2020_CVPR}} &  \multicolumn{1}{c}{$22.4$}  \\
    Non-Semantic      & \multicolumn{1}{l}{AdaIN  \cite{huang17_arbit_style_trans_real_time}}&  \multicolumn{1}{c}{$60.0$}\\
    preservation $s_r$           & MixUp \cite{zhang2017mixup,inoue18_data_augmen_by_pairin_sampl_images_class} & \multicolumn{1}{c}{$00.0$} \\
                            & \multicolumn{1}{l}{SciMix} &   $\mathbf{98.2}$ \\ \bottomrule
  \end{tabular}}
  }
  {\caption{SciMix hybrids properly mix
      semantic and non-semantic contents.}}
\end{figure}

We now verify the auto-encoder described in Sec.~\ref{sec:gen} indeed
yields interesting hybrids for data augmentation. This can be observed
qualitatively by considering the generated hybrids (see Fig.~\ref{fig:exh_sem}) and
quantitatively in Tab.~\ref{tab:exh} through an observation of the generated hybrids in relation to
their parents.

% \paragraph{Generated hybrids}

Fig.~\ref{fig:exh_sem} shows hybrids generated by the method. As can be
observed, SciMix creates hybrids that match the semantic content of the semantic
parent while closely matching the non-semantic parents in every other regard.
Qualitative samples suggest the framework properly identifies semantic content
in the parent samples and successfully embeds the semantic content of the
semantic parent into the non-semantic background of the non-semantic parent.

\paragraph{Preservation of semantic and non-semantic characteristics}

We quantify how well the
generated hybrids $x_h$ inherit properties from the semantic and non-semantic
parents $x_c$ and $x_r$ through the metrics $s_c$ and $s_r$. The semantic transfer rate $s_c$ is the
accuracy of an ``oracle'' classifier (trained on the entire dataset, as a proxy
for human evaluation of hybrid labels) over a
dataset built from hybrids that are assumed to have inherited the label of their
semantic parent. Conversely, the non-semantic preservation rate
$s_r$ is the proportion of hybrids $x_h$ that are closer to the non-semantic
parent $x_r$ in pixel space (ie, $\lVert x_h - x_c \rVert > \lVert x_h - x_r \rVert$).
% As an additional indication, we also provide the average ratio between the
% distances in pixel space: $\frac{\lVert x_h - x_r \rVert}{\lVert x_h - x_c \rVert}$.

Tab.~\ref{tab:exh} shows SciMix compares favorably to texture
altering hybrids (FDA, AdaIN) and standard mixing augmentations (MixUp with the
label of the dominant samples as suggested in
\cite{inoue18_data_augmen_by_pairin_sampl_images_class}) on a strong generator (SVHN 250 labels). While most existing
hybridizations do tend to preserve the semantic content, SciMix shines in that it transfers semantic content while
keeping hybrids very close to their non-semantic parent. Indeed, other hybrids
remain much closer to their semantic parent (always the case - by design - for MixUp).

\subsection{Comparison to Mixing Data Augmentation}

\begin{table}[H]
  \begin{center}
    \caption{Comparison of SciMix with other Data
      Augmentations in low label settings.}
    \label{tab:msda}
    \resizebox{.8\columnwidth}{!}{
      \begin{tabular}{lcc}
        \toprule
        \multirow{2}{*}{Method} & CIFAR10 & SVHN \\ \cmidrule(r){2-2}\cmidrule(r){3-3}
                                & 250 & 60 \\ \midrule
        Mean Teacher, \cite{tarvainen17_mean_teach_are_better_role_model} &   $63.1 \pm 0.9$ & $48.7 \pm 23.0$ \\
        Mean Teacher + MixUp \cite{zhang2017mixup} &  $64.8 \pm 3.5$ & $61.8 \pm 1.0$  \\
        Mean Teacher + CutMix \cite{yun2019cutmix} &  $55.0 \pm 7.5$ & $17.3 \pm 3.7$ \\ \midrule
        SciMix w/ Mean Teacher (ours) & $\mathbf{68 \pm 1.0}$ & $\mathbf{83.4 \pm 0.5}$  \\ \bottomrule
        % Mean Teacher + Hybrids$^*$ (ours) & $61.1 \pm 1.4$ & \\ \bottomrule
      \end{tabular}
    }
  \end{center}
\end{table}

We now show that in very low label settings, the ``artificial'' labelized samples SciMix can outperform
the regularization offered by more traditional mixing Data Augmentation.
Tab.~\ref{tab:msda} shows that SciMix mostly outperforms MixUp and CutMix
for CIFAR10 with 250 labeled samples (the generator is too weak with 100 labels) and SVHN with 60 labeled samples
(hardest setting). While
MixUp does perform similarly to SciMix on CIFAR10, this is likely due to the low
performance of the classifier $E_c \circ C$ trained with only 250 labels.

\subsection{Ablation: Regularization vs. Data Augmentation}

\begin{table}[H]
  \caption{Comparison on the performances of the classifier trained with our generator (Sec.~\ref{sec:gen}) and models trained from scratch without data augmentation
    (Sec.~\ref{sec:da}).}
  \label{tab:regda}
  \begin{center}
    \resizebox{.8\columnwidth}{!}{
    \begin{tabular}{lc}
      \toprule
      \multirow{2}{*}{Method} & \multicolumn{1}{c}{SVHN} \\ \cmidrule(r){2-2}
                              & 60 \\ \midrule
      Mean Teacher, \cite{tarvainen17_mean_teach_are_better_role_model} & $48.7 \pm 23.0$\\
      Generator classifier (Sec.~\ref{sec:gen}) & $72.0 \pm 2.7$  \\
      SciMix w/ Gen. Init. & $72.6 \pm 2.3$  \\
      SciMix w/ Mean Teacher (Sec.~\ref{sec:da}) & $\mathbf{83.4 \pm .5}$ \\ \bottomrule
    \end{tabular}
  }
  \end{center}
\end{table}

While we evaluate our framework as a general data augmentation framework by
using mixed samples from a pre-trained generator to train a model from scratch,
it is interesting to note that our mixing auto-encoder also trains and uses a
classifier model. We study in Tab.~\ref{tab:regda} the relative performance of the generator
classifier (tied to the auto-encoder and regularized by $\mathcal{L}_{hyb}$ but
without mixed data augmentation) and of a classifier trained from scratch with
our mixed data augmentation (but without explicit loss regularization). While the generator classifier does outperform the backbone
classifier in all cases (showcasing the usefulness of the regularization
scheme), training from scratch with data augmentation leads to better results on
every setting. Interestingly, keeping the weights of the auto-encoder classifier
as initialization to train a model with SciMix mixing does not necessarily
outperform a random initialization. This can be explained by the fact that
the separation characterized by the data augmentation is already learned by the
auto-encoder classifier and therefore training with our hybrids fails to
teach the model anything interesting.

\subsection{Model analysis: Importance of the learning scheme}
\label{sec:variants}

\begin{table}[H]
  \begin{center}
    \caption{Comparison of various architectural variants (Sec.~\ref{sec:gen})
      along with samples of generated hybrids for each variant on SVHN 60 labels.}
    \label{tab:variants}

    Hybrids parent pairs $x_r/x_c$ \hspace{1.75cm} 
    \includegraphics[width=.3\columnwidth]{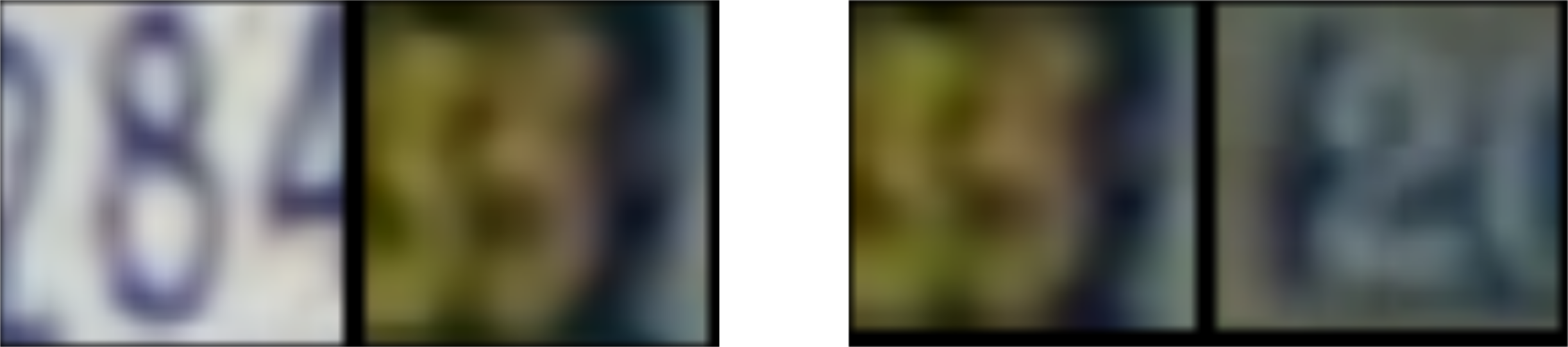} \hfill
    \resizebox{.9\columnwidth}{!}{
      \begin{tabular}{lccccc}
        \toprule
        Method & Accuracy & $s_c$ & $s_r$ & sample hybrids\\ \cmidrule{1-4}
        % Baseline & $48.7 \pm 23.0$ & - & - & - \\ \cmidrule(r){1-4}
        Structural $z_c$ & $39.5 \pm 9.3$ & $16.0$ & $56.0$ & \includegraphics[width=0.2\columnwidth]{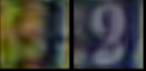}\\
        % Spatial $z_c$ &  $87.4 \pm 3.5$ & $83.7$ & $98.7$ & \includegraphics[width=0.2\columnwidth]{sample_hybrids_spatzc}  \\ \cmidrule(r){1-4}
        % Standard $G$ & \\ \cmidrule(r){1-4}
        No $\mathcal{L}_{hyb}$ & $48.5\pm 14.2$ & $11.7$ & $99.8$ & \includegraphics[width=0.2\columnwidth]{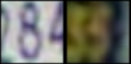} \\
        Basic $\mathcal{L}_{hyb}$ & $66.3 \pm 1.0$ & $66.2$ & $99.4$ & \includegraphics[width=0.2\columnwidth]{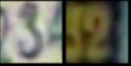}\\
        Non Frozen criterion $\mathcal{L}_{hyb}$ & $81.8 \pm 3.0$ & $75.1$ & $73.8$ & \includegraphics[width=0.2\columnwidth]{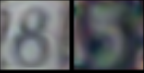}\\ \cmidrule(r){1-4}
        Full Setup & $83.4 \pm 0.5$ & $76.7$ & $98.8$ & \includegraphics[width=0.2\columnwidth]{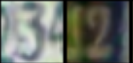}\\ \bottomrule
      \end{tabular}
    }
  \end{center}
\end{table}

To better explore how our framework facilitates the incrustation of semantic
content in the general context of existing samples, we first propose a rapid
ablation study on the quality of the samples generated by variants of our
auto-encoder on our hardest SVHN setting (60 labels). We evaluate this through the classification accuracy of models trained with
the generator's mixed samples, the semantic transfer
$s_c$, the non-semantic transfer $s_r$, and a visual evaluation of two hybrids
(e.g. the leftmost hybrid mixes a blue 8 $x_c$ with a yellow 3 $x_r$).

We consider 4 variations on SciMix's generator to demonstrate the merits of our chosen method:
% \textbf{Architecture: Standard decoder.} We choose to separate the roles of the
% semantic and non-semantic information to ensure the auto-encoder properly
% separates them. We verify this is necessary by considering a standard
% deconvolutional decoder which simply takes the concatenated (unpooled) semantic
% features and (unprojected) non-semantic features as input.
\textbf{$z_r$ as a style code.} We flip the roles of $z_c$ and
$z_r$, to demonstrate $z_c$ is better used as a style code in SciMix.
% \textbf{Architecture: Spatial information in $z_c$.} We treat the semantic
% content as a completely global information (the vector is global average pooled
% from semantic features). To check whether this leads to significant information
% loss, we consider a baseline where the semantic style code retains spatial
% information (flattened instead of pooled).
\textbf{No $\mathcal{L}_{hyb}$.} We train without an explicit
optimization loss, to show $\mathcal{L}_S$ and $\mathcal{L}_{rec}$ are not
sufficient to create good hybrids in SciMix.
\textbf{Basic $\mathcal{L}_{hyb}$.}
We demonstrate the orthogonalization losses $\mathcal{L}_{hyb, class}^-$ and
$\mathcal{L}_{hyb, cont}^-$ contribute to the generator by considering a variant
that does not optimize them.
\textbf{No frozen criterion $\mathcal{L}_{hyb}$.}
We do not force the generator to only optimize the generation of the hybrids
when optimizing $\mathcal{L}_{hyb}$ (the projection of the hybrids in the latent
spaces is also modified).

Tab.~\ref{tab:variants} shows that without explicit hybridization optimization, the model
fails to properly transfer semantic characteristics (low $s_c$ score). Since the
model does properly transfer semantic content with only
$\mathcal{L}_{hybrids}^+$, the addition of the orthogonalization constraints
$\mathcal{L}_{hybrids}^-$ is not necessary to obtain useful hybrids. However,
this orthogonalization increases the diversity in the generated hybrids and
therefore leads to a better augmentation procedure. Not freezing the projection
heads when training for hybridization on the other hand leads to a general
deterioration of the training process and can therefore be felt in both transfer
rates. Predictably, using $z_r$ as a global style code leads a very poor
correspondence of hybrids to their non-semantic parents as things like
backgrounds can become very complicated to reproduce with a modulation based generator.

\subsection{Model analysis: Leveraging the hybrids as Data Augmentation}
\label{sec:loss}

\begin{table}
    \caption{Comparison of various losses to leverage hybrids in $\mathcal{L}_{SciMix}$ (Sec.~\ref{sec:da}) on SVHN 60 labels.}
    \label{tab:loss}
    \centering
    \resizebox{.5\columnwidth}{!}{
      \begin{tabular}{lc}
        \toprule
        Method & Accuracy\\ \midrule
        Baseline &  $48.7 \pm 23.0$ \\
        $\mathcal{L}_{labeled}$  & $29.9 \pm 17.6$ \\
        $\mathcal{L}_{pseudo-label}$  & $62.8 \pm 5.4$ \\
        $\mathcal{L}_{consistency}$ & $77.8 \pm 4.8$ \\
        $\mathcal{L}_{contradict}$  &  $\mathbf{83.4 \pm 0.4}$ \\ \bottomrule
      \end{tabular}
    }
\end{table}

We considered 4 alternative methods to exploit the hybrids generated by our method:
\textbf{Labeled} Only hybrids with a labeled semantic parent are
  considered (supervised training with a hard label)
\textbf{Pseudo-label} Hybrids are treated as labeled samples (hard
  labels), with the labels inherited from the semantic parent's pseudo-labels.
\textbf{Consistency} Hybrids are made to follow their semantic parent's prediction.
\textbf{Contradict} Hybrids are made to match both their semantic
  parent's consistency target and their non-semantic parent targets.

Tab.~\ref{tab:loss} shows that the best results are obtained with
$\mathcal{L}_{contradict}$, but $\mathcal{L}_{pseudo-label}$ and
$\mathcal{L}_{cons}$ also outperform the
baseline method on SVHN 60 labels (hardest setting). The fact
$\mathcal{L}_{contradict}$ outperforms other methods is
interesting in that the loss does not actually treat the hybrids as pure labeled
samples, but assumes some semantic content/noise is retained from the non-semantic
samples. $\mathcal{L}_{labeled}$'s failure suggests
that even with proper mixing, it not possible to improve models by only
generalizing a few labeled samples.

\subsection{Pushing SciMix on CUB-200}
\label{sec:cub}

We now push SciMix further by studying
versions of the more complex \textbf{Caltech-UCSD Birds 200} (CUB-200) dataset
\cite{WelinderEtal2010} (6033 pictures of
200 bird species). Given CUB-200 inherently presents few labels, we directly
study how fully supervised training benefits from SciMix on low labels settings
($\mathcal{L}_S$ is a standard cross-entropy loss).
Furthermore, we take advantage of CUB-200's higher native resolution to go beyond
the limitations of $32\times 32$ images in CIFAR 10 and SVHN: we use SciMix on
commonly studied input sizes $64\times 64$ (e.g. Tiny ImageNet
\cite{chrabaszcz2017}) and $96\times 96$
(e.g. STL-10 \cite{pmlr-v15-coates11a}).

% We furthermore take advantage of CUB-200's higher native resolution to
% demonstrate SciMix's ability to generate convincing hybrids for higher
% dimensional inputs ($64\times 64$ and $96\times 96$ namely).

% We now show SciMix can scale up to more complex tasks by considering the
% \textbf{Caltech-UCSD Birds 200} (CUB-200) dataset \cite{WelinderEtal2010}, composed of 6033 training samples of
% high resolution pictures of 200 bird species. Beyond showcasing the framework's
% ability to model a fully supervised 200 class problem, our study shows SciMix can learn to
% generate convincing hybrids for higher dimensional inputs (resized to $64\times
% 64$ and $96\times 96$ namely).

\begin{figure}
  \subfloat[Examples of SciMix hybrids on CUB $64\times 64$ (left) and
  $96\times 96$ (right).\label{fig:cub}]{
    \begin{minipage}{.11\linewidth}
      {\includegraphics[width=\linewidth]{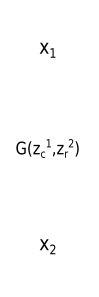}}
    \end{minipage}
    \begin{minipage}{.275\linewidth}
      {\includegraphics[width=\linewidth]{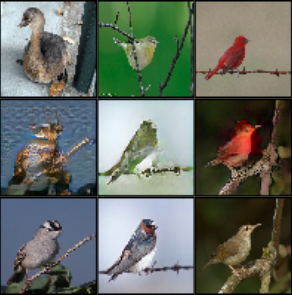}}
    \end{minipage}
    \begin{minipage}{.275\linewidth}
      {\includegraphics[width=\linewidth]{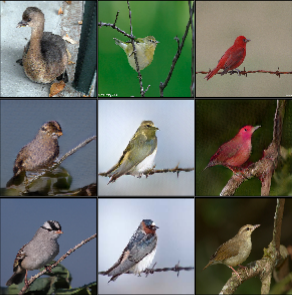}}
    \end{minipage}
  }
  \hfill
  \subfloat[Semantic (resp. non-semantic) transfer rate in hybrids.\label{tab:cub_met}]{
    \resizebox{.24\columnwidth}{!}{
  \begin{tabular}{ccc}
    \toprule
    CUB & $s_c$ & $s_r$ \\ \midrule
    Random & $0.5$ & $50.0$ \\ \midrule
    $64$  &   & \\
    $\times$  &  $43.5$  & $99.8$ \\
    $64$  &    & \\ \midrule
    $96$  &   & \\
    $\times$  &  $41.6$  & $99.8$ \\
    $96$  &    & \\ \bottomrule
  \end{tabular}}
  }
  {\caption{SciMix hybrids mix
      semantic and non-semantic contents on CUB-200.}}
\end{figure}

\paragraph{Quality of generated hybrids}

As can be observed on Fig.~\ref{fig:cub}, the SciMix autoencoder learns to
generate interesting hybrids for different resolutions of the CUB-200 dataset.
While the lower resolution $64 \times 64$ hybrids inherit more semantic
characteristics of the relevant parent, $96 \times 96$ hybrids still retain
semantic patterns tied to the semantic parent's class. 

\begin{table}[H]
  \begin{center}
    \caption{Impact of SciMix data augmentation for
      the CUB-200 dataset at
      resolutions $64\times 64$ and $96\times 96$.}
    \label{tab:cub}
    \resizebox{.8\columnwidth}{!}{
      \begin{tabular}{lcc}
        \toprule
        \multirow{2}{*}{Method} & \multicolumn{2}{c}{CUB-200} \\ 
                                & $64\times 64$ & $96\times 96$ \\ \midrule
        Supervised & $58.9 \pm 1.0$  & $65.2 \pm 0.8$  \\
        SciMix w/ Supervised & $\mathbf{60.2 \pm 0.6}$ & $\mathbf{65.6 \pm 0.9}$ \\ \midrule
      \end{tabular}
    }
  \end{center}
\end{table}

Interestingly, SciMix has no difficulty producing hybrids close to
their non-semantic parent as can be shown in Tab.~\ref{tab:cub_met} by reprising
the analysis of the non-semantic transfer rate $s_r$ from Sec.~\ref{sec:exh}.
Analyzing the semantic transfer rate $s_c$ proves more difficult as our best
``oracle'' classifiers remain unreliable (around $60\%$ accuracy). Nevertheless,
the semantic transfer rates $s_c$ in Tab.~\ref{tab:cub_met} indicate hybrids
generated by SciMix are properly classified by the oracle classifier as having
inherited their semantic parent's class about $40\%$ of the time (orders of
magnitude more than attributable to random chance).

\paragraph{Performance gains}

Tab.~\ref{tab:cub} shows a fully supervised version of
SciMix data augmentation improves supervised models on both $64\times64$ and
$96\times96$ versions of CUB-200. This demonstrates that while SciMix
augmentation strongly benefits from a large amount of unlabeled data,
it can still generate hybrids diverse enough to benefit training with only a
small set of data to generate hybrids from.
\vfill

\section{Conclusion}

In conclusion, we propose in this paper a new approach to mixing data
augmentation that mixes the semantic content of one sample and the non-semantic
content of the other. Making use of advances in style transfer and modular image
generation, we train an auto-encoder that learns to extract and combine semantic
and non-semantic content from multiple images. Through extensive experiments, we
show the intricate hybridization loss we propose leads to the generation of
interesting mixed samples. We furthermore demonstrate it is better to
use semantic information as an indirect ``style code'' input to our StyleGan decoder instead of
non-semantic information.
Afterwards, we prove such
data augmentation significantly improves the training of supervised and semi-supervised models when few
labels are available.

\newpage


%\section*{Acknowledgment}




\bibliographystyle{IEEEtran}
\bibliography{egbib}


% \end{thebibliography}

\end{document}